\documentclass[letterpaper]{article} 
\usepackage{aaai24}  
\usepackage{times}  
\usepackage{helvet}  
\usepackage{courier}  
\usepackage[hyphens]{url}  
\usepackage{graphicx} 
\usepackage{natbib}  
\usepackage{caption} 
\usepackage{algorithm}
\usepackage{algorithmic}
\usepackage{cancel}
\usepackage{amsthm}
\newtheorem*{proposition}{Proposition}
\usepackage{newfloat}
\usepackage{listings}
\DeclareCaptionStyle{ruled}{labelfont=normalfont,labelsep=colon,strut=off} 
\floatstyle{ruled}
\newfloat{listing}{tb}{lst}{}
\floatname{listing}{Listing}
\usepackage{stmaryrd}
\usepackage{amsmath}
\usepackage{amsfonts}
\usepackage{appendix}
\usepackage{bm}
\nocopyright

\title{Variable Importance in High-Dimensional Settings Requires Grouping}
\author{
    Ahmad Chamma\textsuperscript{\rm 1},
    Bertrand Thirion\textsuperscript{\rm 1},
    Denis Engemann\textsuperscript{\rm 2}
}
\affiliations{
    \textsuperscript{\rm 1} Inria, Universite Paris Saclay, CEA \\
    \textsuperscript{\rm 2} Roche Pharma Research and Early Development,
    Neuroscience and Rare Diseases, Roche Innovation Center Basel, F.
    Hoffmann La Roche Ltd., Basel, Switzerland \\
    ahmad.chamma@inria.fr, bertrand.thirion@inria.fr, denis.engemann@roche.com
}

\begin{document}

\maketitle

\begin{abstract}
Explaining the decision process of machine learning algorithms is
nowadays crucial for both a model's performance enhancement and human
comprehension.
This can be achieved by assessing the variable importance of single
variables, even for high-capacity non-linear methods, e.g.
Deep Neural Networks (DNNs).
While only removal-based approaches, such as Permutation Importance
(\textit{PI}), can bring statistical validity, they return misleading
results when variables are correlated.
Conditional Permutation Importance (\textit{CPI}) bypasses PI's
limitations in such cases.
However, in high-dimensional settings, where high correlations between
the variables cancel their conditional importance, the use of
\textit{CPI} as well as other methods leads to unreliable results,
besides prohibitive computation costs.
Grouping variables statistically via clustering or some prior
knowledge gains some power back and leads to better interpretations.
In this work, we introduce \textit{BCPI} (Block-Based Conditional Permutation
Importance), a new generic framework for variable importance computation with
statistical guarantees handling both single and group cases.
Furthermore, as handling groups with high cardinality (such as a set
of observations of a given modality) are both time-consuming and
resource-intensive, we also introduce a new stacking approach
extending the DNN architecture with sub-linear layers adapted to the
group structure.
We show that the ensuing approach extended with stacking controls the
type-I error even with highly-correlated groups and shows top accuracy
across benchmarks.
Furthermore, we perform a real-world data analysis in a large-scale medical
dataset where we aim to show the consistency between our results and the
literature for a biomarker prediction.
\end{abstract}

\section{Introduction}

Machine Learning (ML) algorithms are extensively used in many fields of science, such as biomedical application
\cite{strzeleckiMachineLearningBiomedical2022, alberIntegratingMachineLearning2019},
neuroscience~\cite{koraEEGBasedInterpretation2021, knutsonIntegratingBrainImaging2020},
and social sciences \cite{lundbergResearcherReasoningMeets2022,chenSocialPredictionNew2021}.
The increasing importance of ML in society raises issues of accountability, hence, stimulating research on interpretable ML.
Reaching a comprehensive understanding of the decision process is crucial for providing statistical and, ideally, scientific insights to the practitioner \cite{gaoLazyEstimationVariable2022,
molnarModelagnosticFeatureImportance2021, flemingHowWhyInterpret2020,
hookerBenchmarkInterpretabilityMethods2019}.

To gauge the impact of variables on model prediction, aka
\emph{variable importance}, several model-agnostic attempts have emerged
\cite{molnarInterpretableMachineLearning2022, ribeiroWhyShouldTrust2016a}.
Examples include \textit{Permutation Feature Importance (PFI)}
\cite{breimanRandomForests2001}, \textit{Conditional Randomization Test}
\cite{candesPanningGoldModelX2017} and \textit{Leave-One-Covariate-Out (LOCO)}
\cite{leiDistributionFreePredictiveInference2018}.
All these instances constitute removal-based approaches \cite{covertUnderstandingGlobalFeature2020}, and are so far, the only ones known to provide statistically grounded measures of significance.
Importantly, removal-based approaches require retraining the model
after removing the variable of interest and are, therefore, time-consuming.
Moreover, the common Permutation Importance (\textit{PI}, \citealt{breimanRandomForests2001}) risks mistaking insignificant variables for significant ones when variables are correlated \cite{hookerUnrestrictedPermutationForces2021}.
Conditional Permutation Importance
\textit{CPI} can overcome these limitations
\cite{bleschConditionalFeatureImportance2023, watsonTestingConditionalIndependence2021,
debeerConditionalPermutationImportance2020, fisherAllModelsAre2019,chammaStatisticallyValidVariable2023}.
However, in high-dimensional settings, single variable importance
computation suffers from very high correlation between the variables
\cite{chevalierDecodingConfidenceStatistical2021}.
More precisely, this makes conditional importance estimation less
informative, as it remains unclear how much information each
variable adds.
In the extreme case where variables are duplicated, conditional
importance can no longer be defined.
More generally, correlations larger than $.8$ are known to present a
hard challenge, at least for linear learners
\cite{chevalierDecodingConfidenceStatistical2021}.
Importance analysis then typically yields spuriously significant variables, which ruins its ability to statistically control the false positive rate \cite{stroblConditionalVariableImportance2008}.
Besides, examining the importance of each of the hundreds or thousands variables separately will result in prohibitive computation costs \cite{covertUnderstandingGlobalFeature2020} ---removal procedures typically have quadratic complexity--- and defy model interpretability.

Group-based analysis can offer a remedy at it regularizes power estimates and leads to reduced computation time \cite{molnarGeneralPitfallsModelAgnostic2021, buhlmannStatisticalSignificanceHighdimensional2013}.
This can improve inference as it helps handle the
curse of correlated variables in high-dimensional settings.
So far, common group-based methods neglected investigating statistical guarantees, in particular, type-I error control, i.e. the percentage of irrelevant variables identified as relevant (false positives).
Statistical error control for groups obviously requires information on variable grouping available through two strategies: \textit{Knowledge-driven} grouping, where
the variables are grouped based on their domain-specific information rather than their shared statistical properties and \textit{Data-driven} grouping, where clustering approaches are used such as hierarchical or divisive clustering.

Grouping has also been successfully performed for multimodal applications
~\cite{albuMMStackEnsNewDeep2023,
engemannCombiningMagnetoencephalographyMagnetic2020,
rahimIntegratingMultimodalPriors2015} via model stacking \cite{wolpertStackedGeneralization1992} which is typically based on pipelines of disconnected models.

\paragraph{Contributions} We propose \textit{Block-Based
Conditional Permutation Importance (BCPI)}, a new framework for variable
importance computation (single and group levels) with explicit statistical guarantees (p-values).

\begin{itemize}
    
\item Following our review of the literature (section~\ref{related_work}),
we provide theoretical results on group-based conditional permutation importance
(section~\ref{group_conditional}).
\item We propose a novel \textit{internal stacking} approach by extending the architecture of our default Deep Neural Network (DNN) model with the use of a linear projection of the groups, which can significantly reduce computation time (section~\ref{stacking_sec}).
\item We conduct extensive benchmarks on synthetic and
real world data (section \ref{experiments}) which demonstrate the capacity of the proposed method to combine high prediction performance with theoretically grounded identification of predicatively important groups of variables.
\item We provide publicly available code (compatible with the Scikit-learn API) on GitHub (\url{https://github.com/achamma723/Group_Variable_Importance}).
\end{itemize}

\section{Related work}
\label{related_work}

Group-based variable importance has been introduced for Random Forests by \cite{wehenkelRandomForestsBased2018}, extending the seminal work of \citet{louppeUnderstandingVariableImportances2013} on \textit{Mean Decrease Impurity (MDI)}.
Once all the variables have their corresponding impurity function scores, the
importance score of the group of interest are ($1$) the sum, ($2$) the
average or ($3$) the maximum of the impurity scores among the participating
variables.
Despite that, ($1$) the sum displays bias in favor of larger-sized groups, ($2$)
the average diminishes a group's significance when only a small fraction of its
features hold importance and ($3$) the maximum suggests that the sole most important feature reflects the collective importance of the group.

\citet{williamsonGeneralFrameworkInference2021} proposed a
model-agnostic approach based on refitting the learner after the removal of a variable of interest also called \textit{LOCO
(Leave-One-Covariate-Out)} by \citet{leiDistributionFreePredictiveInference2018}.
This work has then been adapted to the group-level by considering the removal
of all the variables of the group of interest jointly, as in
\textit{Leave-One-Group-Out (LOGO)} presented in \cite{auGroupedFeatureImportance2021}.
In lieu of removing the group of interest,
\citet{auGroupedFeatureImportance2021} established \textit{Leave-One-Group-In
(LOGI)} that assesses the impact of the group of interest on the
prediction compared to the null model - the prediction is the average of the outcome.
However, this approach becomes intractable easily due to the necessity of refitting the
learner for each group, particularly in the case of low cardinality groups.

\citet{miPermutationbasedIdentificationImportant2021} proposed an efficient
model-agnostic procedure for black-box models' interpretation.
It uses the \textit{permutation approach} \cite{breimanRandomForests2001,
fisherAllModelsAre2019} with the importance score computed as the reduction in a
model's performance when randomly shuffling the variable of interest.
To account for group-level structure,
\cite{gregoruttiGroupedVariableImportance2015} suggested taking into
account all the variables of the group of interest in the permutation
scheme jointly, known as \textit{Group Permutation Feature Importance
    (GPFI)}.
\citet{auGroupedFeatureImportance2021} proposed \textit{Group Only
Permutation Feature Importance (GOPFI)} which examines the level of the group's
individual contribution to the model's performance.
The random joint shuffling is performed for all the variables of the
different groups expect the ones of the group of interest.
However, according to \citet{stroblConditionalVariableImportance2008},
simple permutation approaches yield poor accuracy and specificity in
high correlation settings.
\citet{leeUnderstandingLearnedModels2018} applied perturbations to the variables
and groups of interest while providing p-values.
Nevertheless, they did not focus on the degree of correlation between the
variables (and the groups) which increases the difficulty of the problem. 

A different angle can be motivated by a recent line of work that developed model-stacking techniques \cite{wolpertStackedGeneralization1992} which combine different input domains and groups of variables rather than aggregating different estimators on the input data. 
This approach has been used in various applications ranging from video analysis
\cite{zhouMultimodalFeatureFusion2021} over protein-protein interactions
\cite{albuMMStackEnsNewDeep2023} to neuroscience applications
\cite{rahimIntegratingMultimodalPriors2015}.
A key benefit of multimodal or group stacking  is that it allows for modality-specific encoding strategies and while approaching inference at the simplified level of the 2\textsuperscript{nd} level model combining the modality-wise predictions or activations.
This strategy has been used to explore importance of distinct types of brain activity
at different frequencies for age prediction \cite{sabbaghRepurposingElectroencephalogramMonitoring2023,
engemannCombiningMagnetoencephalographyMagnetic2020}.
While stacking  is easy to implement with standard software e.g. scikit-learn \cite{pedregosaScikitlearnMachineLearning2011}, inference with stacking has not been formalized yet. Moreover, it requires fitting multiple disconnected estimators which may limit the capacity of the model.

\section{BCPI and \textit{Internal Stacking} Approach}
\label{bbi-stack}
\begin{figure*}[t]
    \centering
    \includegraphics[width=\textwidth]{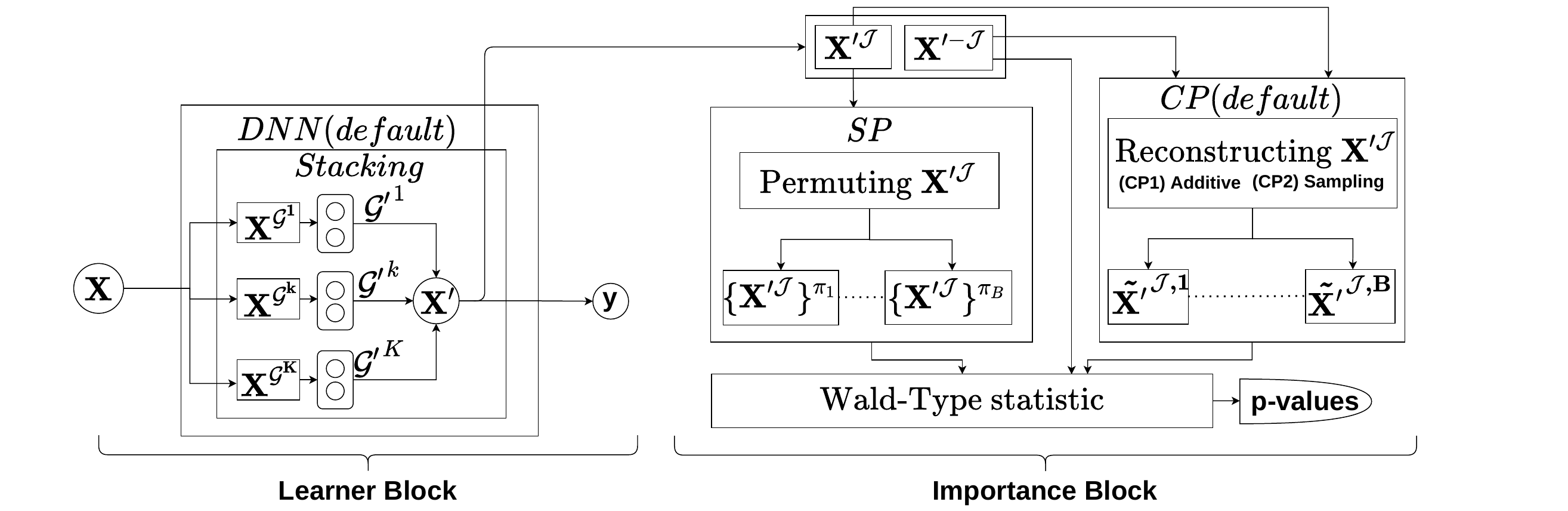}
    \caption{\textbf{Block-Based Conditional Permutation Importance}: Framework
    for single/group variable importance computation with statistical
    guarantees. \textbf{(Learner Block)} The learner used to predict the
    outcome $\mathbf{y}$ from the design matrix $\mathbf{X}$. \textit{Internal stacking}  linearly projects each group by the mean of an extra linear sub-layer. \textbf{(Importance Block)}:
    Reconstruction of the  group of interest $\mathbf{X'^{\mathcal{J}}}$ is
    accomplished via \textit{CP (Conditional Permutation)} block with
    (\textbf{CP1}) the additive or (\textbf{CP2}) the sampling constructions as
    stated in section~\ref{group_conditional}. The permutation scheme can be changed to standard permutation (SP).}
    \label{stacking}
\end{figure*}
\subsection{Preliminaries}
\paragraph{Notations}
We denote by matrices, vectors, scalar variables and sets by bold uppercase
letters, bold lowercase letters, script lowercase letters, and calligraphic
letters, respectively (e.g. $\mathbf{X}$,
$\mathbf{x}$, $x$, $\mathcal{X}$).
Designating by $\mu$ the function that maps the sample space
$\mathcal{X} \subset \mathbb{R}^{p}$ to the outcome space $\mathcal{Y}
\subset \mathbb{R}$ and $\hat{\mu}$ is an estimate of $\mu$ within a
certain class $\mathcal{F}$ of estimators.
We express by $\llbracket n \rrbracket$ the set \{1, \dots, $n$\}, by $\langle
., .\rangle$ the standard dot product and by ($\pi$) the shuffling process. \\ \\
Let $\mathcal{S} = \{ \mathcal{G}^{k}, k \in \llbracket K \rrbracket \}$ and $\mathcal{S'} = \{\mathcal{G'}^{k}, k \in \llbracket K \rrbracket\}$
be  the set of $K$
pre-defined subset of variables in the data and the set of $K$ new
subset of variables following linear projections with a set
$\mathcal{P}$ of projection matrices, respectively.
Projection matrices are meant to produce a group summary of the information.
Let $\mathcal{P} = \{\mathbf{U_{k}}, k \in \llbracket K \rrbracket \}$ be the set of projection matrices $\mathbf{U_{k}} \in \mathbb{R}^{|\mathcal{G}^{k}| \times |\mathcal{G'}^{k}|}$.
Let $\mathcal{J} = \{j_{1}, \dots, j_{r} \} \in (\mathcal{S} \cup \mathcal{S'})$
be a subset of $r$ variables with consecutive indices in $\llbracket p
\rrbracket$, $r \leq p$.
Let $\mathbf{X} \in \mathbb{R}^{n \times p}$ be a design matrix where $i^{th}$
row, $j^{th}$ column and $\mathcal{J}^{th}$ subset of columns are
indicated by $\mathbf{x_{i}}$, $\mathbf{x^{j}}$ and $\mathbf{X^{\mathcal{J}}}$
respectively.
Let $\mathbf{X^{\mathcal{-J}}} = (\mathbf{x^{1}}, \dots, \mathbf{x^{j_{1} -
    1}},
    \mathbf{x^{j_{r} + 1}}, \dots, \mathbf{x^{p}})$ be the design
matrix with the $\mathcal{J}^{th}$ subset of variables is removed.
Let $\mathbf{X^{\mathcal{(J)}}} = (\mathbf{x^{1}}, \dots, \mathbf{x^{j_{1} -
    1}}, \{ \mathbf{x^{j_{1}}} \}^{\pi}, \dots,
\{ \mathbf{x^{j_{r}}} \}^{\pi}, \dots, \mathbf{x^{p}})$ be the design
matrix with the $\mathcal{J}^{th}$ subset of variables is shuffled.
The rows of $\mathbf{X^{\mathcal{-J}}}$ and $\mathbf{X^{(\mathcal{J})}}$ are
denoted $\mathbf{x_{i}^{\mathcal{-J}}}$ and $\mathbf{x_{i}^{\mathcal{(J)}}}$
respectively, for $i \in \llbracket n \rrbracket$.
Let $\mathbf{X'}$ be the linearly projected version of
$\mathbf{X}$ via $\mathcal{P}$ where $p' = \sum_{k=1}^{K} |\mathcal{G'}^{k}|$.
\paragraph{Problem Setting} 
We consider the regression or the classification problem where the
response vector $\mathbf{y} \in \mathbb{R}^{n}$ or $\in \{0, 1\}^{n}$ respectively
and the design matrix $\mathbf{X} \in \mathbb{R}^{n \times p}$ (encompasses $n$
observations of $p$ variables), along with $\mathcal{S}$ (i.e. $K$ pre-defined groups).
Across the paper, we rely on an i.i.d. sampling train/validation/test partition
scheme where the $n$ samples are divided into $n_{train}$ training and
$n_{test}$ test samples.
The train samples were used to train $\hat{\mu}$ with empirical risk minimization.
This function is utilized for appraising the importance of variables on a novel
dataset (test set).

\subsection{Group conditional variable importance}
\label{group_conditional}
We define the joint permutation of group
$\mathbf{x}^\mathcal{J}$ conditional to $\mathbf{x^{-\mathcal{J}}}$, as a
group $\mathbf{\tilde{x}^{\mathcal{J}}}$ that preserves the joint dependency of
$\mathbf{x^{\mathcal{J}}}$ with respect to the other variables in
$\mathbf{x^{-\mathcal{J}}}$, although the independent part is shuffled.
The reconstruction of $\mathbf{\tilde{x}^{\mathcal{J}}}$ is done via two
approaches, both, based on fast approximation with a lean model: (\textbf{1}) Additive construction combines the prediction
of a Random Forest using the remaining groups and a shuffled version of the residuals i.e. $\mathbf{\tilde{x'}}^{\mathcal{J}} =
\mathbb{E}(\mathbf{{x'}^{\mathcal{J}}}|\mathbf{{x'}^{\mathcal{-J}}}) +
(\mathbf{{x'}^{\mathcal{J}}} -
\mathbb{E}(\mathbf{{x'}^{\mathcal{J}}}|\mathbf{{x'}^{\mathcal{-J}}}))^{\pi}$
where the residuals of the regression of $\mathbf{{x'}^{\mathcal{J}}}$ on
$\mathbf{{x'}^{\mathcal{-J}}}$ are shuffled. \textbf{(2)} Sampling construction uses a Random Forest model to fit
$\mathbf{{x'}^{\mathcal{J}}}$ from $\mathbf{{x'}^{\mathcal{-J}}}$,
followed by sampling the prediction from within its leaves.
When dealing with regression, this results in the following importance estimator:
\begin{equation}
    \hat{m}^{\mathcal{J}}_{CPI} = \frac{1}{n_{test}} \sum_{i=1}^{n_{test}} \left( (y_i - \hat{\mu}(\mathbf{\tilde{x}_i^{(\mathcal{J})}}))^2 - (y_i - \hat{\mu}(\mathbf{x_i}) )^2 \right),
    \label{eq:cpi}
\end{equation}
where $\mathbf{\tilde{X}^{(\mathcal{J})}} = (\mathbf{x^{1}}, \dots, \mathbf{x^{j_{1}-1}},
\mathbf{\tilde{x}^{j_{1}}}, \dots,
\mathbf{\tilde{x}^{j_{r}}}, \\ \dots, \mathbf{x^{p}}) \in
\mathbb{R}^{n_{test} \times p}$ be the new design matrix including the
remodeled version of the group of interest $\mathbf{X^{\mathcal{J}}}$.

In Fig.~\ref{stacking}, we introduce \textit{BCPI} a novel general
framework for variable importance, at both single and group levels,
yielding statistically valid p-values.
It consists of two blocks: a \emph{Learner Block} defined by the prediction model of interest 
\emph{Importance Block} reconstructing the variable (or group) of interest via conditional permutation (CP) --
$\hat{m}^{\mathcal{J}}_{CPI}$.
The implementation provided with this work supports estimators compatible with the scikit-learn API for both blocks.
Yet, our default method \textit{BCPI-DNN} is adapted with: (1) a DNN as a base
learner for its high predictive capacity inspired from
\cite{miPermutationbasedIdentificationImportant2021} and (2) a Random Forest, a
less powerful, but much simpler, yet, still generic model as a conditional probability learner.
For study purposes, the framework is also adapted with the standard permutation
scheme through the (SP) block (labeled \textit{BPI}).
The theoretical results, conditions underlying this proposition as well as limitations of (\textit{PI}) were developed in~\cite{chammaStatisticallyValidVariable2023} and adapted to the group setting (supplementary materials).

\begin{proposition}
    Assuming that the estimator $\hat{\mu}$ is obtained from a class of
    functions $\mathcal{F}$ with sufficient regularity, i.e. that it
    meets conditions of A1: optimality, A2: differentiability, A3: continuity of optimization, A4: Continuity of derivative, B1: Minimum rate of convergence and B2: Limited complexity, the importance score $\hat{m}^{\mathcal{J}}_{CPI}$
    defined in (\ref{eq:cpi}) cancels when $n_{train} \rightarrow
    \infty$ and $n_{test} \rightarrow \infty$ under the null
    hypothesis, i.e. the $\mathcal{J}^{th}$ group is not significant for the prediction. Moreover, the Wald statistic $z^\mathcal{J} =
    \frac{mean(\hat{m}^{\mathcal{J}}_{CPI})}{std(\hat{m}^{\mathcal{J}}_{CPI})}$
    obtained by dividing the mean of the importance score by its standard
    deviation asymptotically follows a standard normal distribution.
\end{proposition}
This implies that in the large sample limit, the p-value associated with
$z^{\mathcal{J}}$ controls the type-I error rate for all optimal estimators in $\mathcal{F}$.
It entails making sure that the importance score defined in (\ref{eq:cpi}) is 0
for the class of learners that meet  specific convergence guarantees and are
immutable to arbitrary change in their $\mathcal{J}^{th}$ arguments, conditional
on the others.
We also state the precise technical conditions under with
$\hat{m}^{\mathcal{J}}_{CPI}$ used is (asymptotically) valid, i.e. leads to a
Wald-type statistic that behaves as a standard normal under the null
hypothesis.
As a result, all terms in Eq.~\ref{eq:cpi} vanish with speed
$\frac{1}{\sqrt{n_{test}}}$ from the \emph{Berry-Essen} theorem, under the
assumption that the test samples are i.i.d.

\subsection{Internal Stacking}
\label{stacking_sec}
The vector $\mathbf{x} \in \mathcal{X}$ is composed of $K$ groups in
$\mathcal{S}$, each considered as an independent input modality.
Performing column slicing on $\mathbf{x}$, according to $\mathcal{S}$, yields
the set $\{ \mathbf{x^{\mathcal{G}^{k}}}, k \in \llbracket K \rrbracket \}$.
A linear transformation to a lower space is applied on each input modality
$\mathbf{x^{\mathcal{G}^{k}}}$ through the set of projection matrices
$\mathcal{P}$ producing a linear variant denoted $\mathbf{x'^{k}}$ as:
\begin{equation*}
    \mathbf{{x'}^{k}} = <\mathbf{x^{\mathcal{G}^{k}}}, \mathbf{U_{k}}>,
\end{equation*}
where $k \in \llbracket K \rrbracket$.

Concatenating the set of linear variants $\{\mathbf{{x'}^{k}}, k \in \llbracket
K \rrbracket \}$ provides the linear version of $\mathbf{x}$ i.e. the vector $\mathbf{x'}$.
If the new space is a unidimensional Euclidean space i.e. $\mathbf{{x'}} \in
\mathbb{R}^{K}$, a group summary of the information within all groups is
returned, and the problem is reduced to the single-level case. 
However, if the new space is not unidimensional, we then have a dimension
reduction, where the group summary of information is exclusive per group
(multioutputs per group).
In this case, the new groups contained in $\mathbf{x}$ are denoted
$\mathcal{G'}^{k}$ with the corresponding linear variant
$\mathbf{x'}^{\mathcal{G'}^{k}}$ as seen in Fig.~\ref{stacking}.
Instead of performing stacking in a separate estimation step under a different
learner, we have incorporated it to the inference process, thus learning a
consistent new presentation of the groups.
This is simply implemented as an initial linear sub-layer without activation in
the $\hat{\mu}$ network.
Therefore, $\mathbf{{x'}^{k}}$ can be seen analogous to the predictions from the input models in a classical stacking pipeline that are forwarded to the meta learner, hence, $\mathbf{{x'}^{k}}$ can be treated like a regular data column by permutation algorithms. 

\section{Experiments}
\label{experiments}
To ensure a fair comparison across experiments, we use all methods
with their original implementation.
As for \textit{BCPI-DNN}, \textit{BCPI-RF} and \textit{BPI-DNN} particularly,
the default behavior consists of a 2-fold internal cross validation where the
importance inference is performed on an unseen test set.
The scores from different splits are thus concatenated to compute the final
variable importance.
All experiments are performed with $100$ runs.

\subsection{Experiment 1: Benchmark of grouping methods}
\label{exp1}
\begin{figure*}[t]
    \centering
    \includegraphics[width=\textwidth]{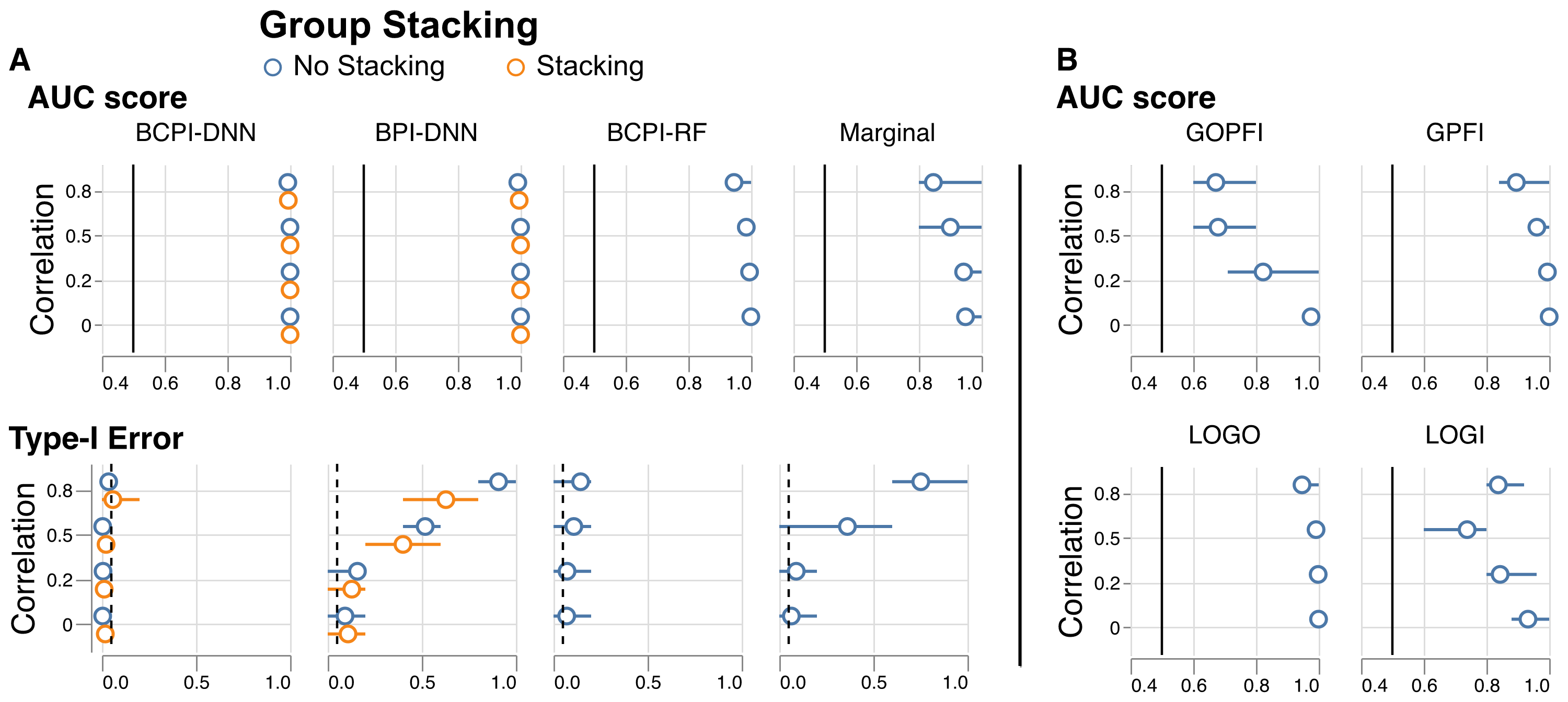}
    \caption{\textbf{Benchmarking grouping methods}: \textit{BCPI-DNN} is
    compared to baseline models and competing approaches for group variable
    importance. \textbf{(A)} AUC score (correct ranking of variables) and Type-I error ($\text{p-val} < 0.05$) for methods providing p-values. \textbf{(B)} AUC scores for methods not providing p-values. Prediction tasks were
    simulated with $n$ = $1000$ and $p$ = $50$. Dashed line: targeted type-I error rate at 5\%. Solid line: chance level.}
    \label{bench_plot}
\end{figure*}
We include \textit{BCPI-DNN} in a benchmark with other state-of-the-art methods
for group-based variable importance.
The data $\{\mathbf{x_{i}}\}_{i=1}^{n}$ follow a Gaussian distribution with a
predefined covariance structure $\mathbf{\Sigma}$ i.e. $\mathbf{x_{i}} \sim
\mathcal{N}(0, \mathbf{\Sigma}) \forall i \in \llbracket n \rrbracket$.
We consider a block-designed covariance matrix $\mathbf{\Sigma}$ of $10$ blocks
with an intra-block correlation coefficient $\rho_{intra} = 0.8$ among the variables
of each block and an inter-block correlation coefficient $\rho_{inter} \in \{0,
0.2, 0.5, 0.8\}$ between the variables of the different blocks.
Each block is considered as a separate group.
In this experiment, $n = 1000$ and $p = 50$ i.e. we have 5 variables per block/group.
We defined an important group as a group having at least one variable that took
part in simulating the outcome $y$.
Thus, to predict $y$, we rely on a linear model where the
first variable of each of the first 5 groups is used in the following model:
\begin{equation}
    y_{i} = \mathbf{x_{i}} \bm{\beta} + \sigma \epsilon_{i}, \forall i \in \llbracket n \rrbracket
\label{linear_model}
\end{equation}
where $\bm{\beta}$ is a vector of regression coefficients having only $5$
non-zero coefficients (the true model), $\bm{\epsilon} \in
\mathcal{N}(0,\mathbf{I})$ is the Gaussian additive noise with magnitude $\sigma
= \frac{||\mathbf{X} \bm{\beta}||_{2}}{SNR \sqrt{n}}$.
We used the same setting from \cite{janitzaComputationallyFastVariable2018}
where the $\bm{\beta}$ values are drawn i.i.d. from the set $\mathcal{B} = \{\pm
3, \pm 2, \pm 1, \pm 0.5 \}$.
We consider the following state-of-the-art baselines:
\begin{itemize}
    \item Marginal Effects: A multivariate linear model is applied to each group separately.
    Importance scores correspond to ensuing p-values.
    \item Leave-One-Group-In (\textit{LOGI})
    \cite{auGroupedFeatureImportance2021}: Similar to \textit{Marginal Effects}
    using a Random Forest. Provides no p-values.
    \item Leave-One-Group-Out (\textit{LOGO}) \cite{williamsonGeneralFrameworkInference2021}: Refitting of
    the model is performed after removing the group of interest.
    \item Group Only Permutation Feature Importance (\textit{GOPFI}) \cite{auGroupedFeatureImportance2021}: Joint permutation of
    all variables except for those of the group of interest.
    \item Group Permutation Feature Importance (\textit{GPFI}) \cite{gregoruttiGroupedVariableImportance2015,
    valentinInterpretingNeuralDecoding2020}: 
    Joint permutation of all variables of the group of interest.
\end{itemize}

\noindent	 
In addition, we benchmarked the three variants of our proposed method:

\begin{itemize}
    \item BPI-DNN: Similar to \textit{GPFI} based on a DNN estimator. It is also enhanced by the new \textit{internal stacking} approach.
    \item BCPI-RF: BCPI where $\hat{\mu}$ is obtaind from a Random Forest.
    \item BCPI-DNN: BCPI where
        $\hat{\mu}$ is a DNN. It is also enhanced by the new \textit{internal stacking} approach.
\end{itemize}
\subsection{Experiment 2: Impact of Stacking}
\label{exp2}
To assess the impact of performing stacking regarding accuracy in inference and computation time, we conducted a comparison restricted to \textit{BCPI-DNN}.
We relied on the same covariance structure setting as in Experiment $1$ with an
intra-block correlation coefficient $\rho_{intra} = 0.8$ and an inter-block
correlation coefficient $\rho_{inter} = 0.8$.
The number of samples $n$ and the number of variables $p$ were both set to $1000$
i.e. the number of variables per block/group increased to 100 in order to build
groups with high cardinality.
The outcome $y$ was simulated using the same model as in
Eq.~\ref{linear_model} where a group is predefined as important having at least
$10\%$ of its variables taking part in computing the outcome.
\begin{figure*}[t]
    \centering
    \includegraphics[width=\textwidth]{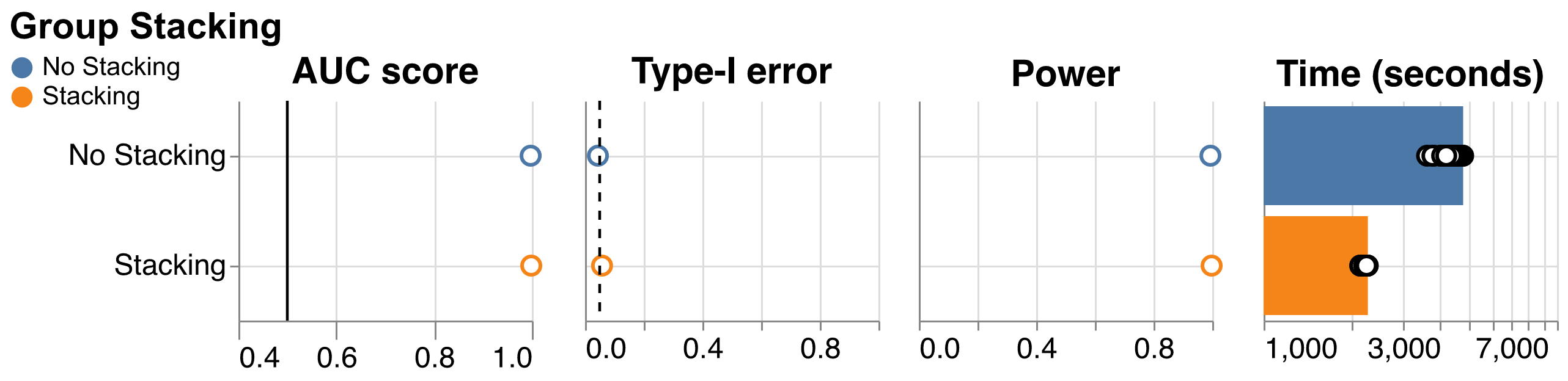}
    \caption{\textbf{Impact of Stacking}: Performance at detecting
    important groups on simulated data with $n$ = $1000$ and $p$ = $1000$ with
    $10$ blocks/groups, each group having a cardinality of $10$. AUC scores and Type-1 error as in Fig.~\ref{bench_plot}. \textbf{(Power)} quantifies the average proportion of detected informative variables ($\text{p-value} <
    0.05$). Panel \textbf{(Time)} displays computation time in seconds with
    $\log_{10}$ scale per core on $100$ cores. Dashed line: targeted type-I error
    rate. Solid line: chance level.}
    \label{compare_plot}
\end{figure*}
\subsection{Experiment 3: Age prediction with UKBB}
\label{exp3}
\begin{figure*}[t]
    \centering
    \includegraphics[width=\textwidth]{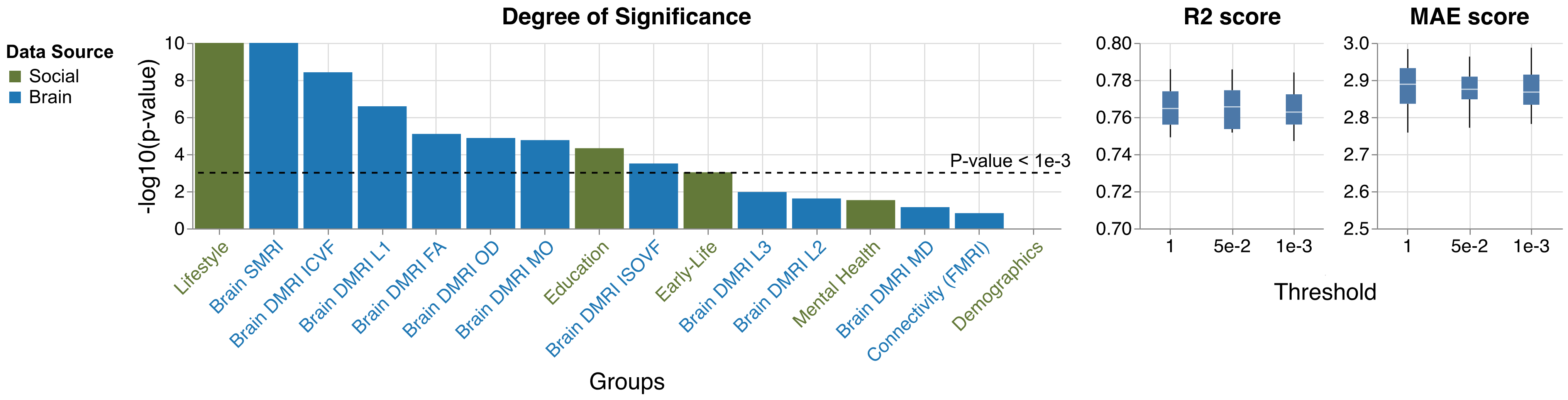}
    \caption{\textbf{Brain Age prediction in UKBB}: Prediction of brain age from
    various socio-demographic and brain-imaging groups of phenotypes in a sample
    of $n$ = $8357$ volunteers from the UK BioBank. (\textbf{Degree of
    significance}) plots the level of significance for the different brain (in
    blue) and social (in red) groups in terms of $-\log_{10}$ of the derived p-values.
    Dashed line: targeted type-I error rate at $p = 0.001$. (\textbf{R2 score \& MAE
    score}) checks the performance of the trained learner when retaining all the
    groups vs removing non-significant groups.}
    \label{ukbb_app}
\end{figure*}
We conducted an empirical benchmark of the performance of
\textit{BCPI-DNN} combined with \textit{internal stacking} in a real-world biomedical dataset.
The UK Biobank project (UKBB) encompasses imaging and socio-demographic derived phenotypes from a prospective cohort of participants drawn from the population of the UK \cite{constantinescuFrameworkResearchContinental2022, littlejohnsUKBiobankImaging2020}.
In the past years, the UKBB dataset has enabled large-scale studies investigating associations between various phenotypes (physiological, cognitive) and environmental or life-style  factor. This has given rise to successful analysis of factors associated to personal well-being and health~\cite{newbyAssociationsBrainVolumes2021, mutzLifetimeDepressionAgerelated2021} at an epidemiological scale.
In the context of machine learning with brain data, age-prediction is an actively studied task which can provide a normative score when applying a reference model on clinical cohorts~\cite{colePredictingAgeUsing2017}.
State-of-the-art models were based on convolutional neural networks and report mean absolute errors between 2-3 years~\cite{roibuBrainAgesDerived2023,jonssonBrainAgePrediction2019}.
Recent extensions have focused on MRI-contrast and region-specific insights,
often based on informal inference~\citep{roibuBrainAgesDerived2023,
popescuLocalBrainAgeUNet2021}.
Another line of work~\cite{dadiPopulationModelingMachine2021,anaturkPredictionBrainAge2021} has focused on other sources of normative ageing information, highlighting cognitive social and lifestyle factors.
In this context, the analysis of importance of multimodal inputs has so far been hampered by the lack of formal inference procedures and the high-dimensional setting with highly correlated variables.

We approached this open task using the proposed method, reusing the pre-defined groups in the
work by~\cite{dadiPopulationModelingMachine2021}. 
We focused on data from participants who attended the
imaging visit ($n$ = $8357$) to study the group-level importance rankings provided by \textit{BCPI-DNN}.
We defined important groups by p-value threshold of $< 10^{-3}$.
While this setting lacks an explicit ground truth for the important groups, we explored the appropriate group selection through model performance in terms of ($R^2$ \& MAE scores, 10-fold cross-validation) after removing the non-significant groups.
We accessed the UKBB data through its controlled access scheme in accordance with its institutional ethics boards~\cite{bycroftUKBiobankResource2018,sudlowUKBiobankOpen2015}.

\section{Results}
\label{results}
We benchmarked state-of-the-art baselines and the proposed methods across data-generating scenarios under increasing
inter-block correlation strength \{0, 0.2, 0.5, 0.8\} (Fig.~\ref{bench_plot}).
\textit{BCPI-DNN} and \textit{BPI-DNN} were implemented in two variants: with or without the novel \textit{internal stacking}. 
For the AUC score, we observed that (\textit{BCPI-DNN} \& \textit{BPI-DNN} -
based on the DNN) and (\textit{BCPI-RF}, \textit{GPFI} \& \textit{LOGO} -
based on Random Forests) showed the highest performance across the
different scenarios, hence, accurately ordering the variables according to their significance.
As expected, the \textit{Marginal} baseline performed lowest as it could not access any conditional information.
\textit{GOPFI} and \textit{LOGI} both suffered when the correlation between the groups increased, which is not surprising.
Considering false positive rate, \textit{BCPI-DNN} controlled the type-I error at the targeted rate (5 \%) while \textit{BPI-DNN}--- based on the standard permutation of the group of interest--- failed to do so in the setting of high correlations between the groups, and thus provided spurious results.
Interestingly, for \textit{BPI-DNN}, \textit{internal stacking}
slightly increased its capacity to control the type-I error.
\textit{BCPI-RF}--- based on the conditional importance with Random
Forests--- better controlled the type-I error compared to \textit{BPI-DNN}.
Nevertheless, in the presence of strong correlations, it did not fully reach the target rate.
Additional analyses suggested that the \textit{marginal} approach failed in the current setting, whereas on average, the DNN had higher scores ($R^2 \sim 0.95$) than the Random Forest ($R^2 \sim 0.8$).
Additional analyses of performance in terms of power and computation time of showed that \textit{BCPI-DNN}, \textit{BPI-DNN}, \textit{BCPI-RF} and \textit{Marginal} showed favorable results compared to other baselines and competing methods.
%

The AUC score, type-I error, power and computation time for Experiment~\ref{exp2} are presented in Fig.~\ref{compare_plot}.
\textit{BCPI-DNN} with \textit{internal stacking} performed similarly as the same approach without stacking.
Thus, both approaches showed comparable inferential behavior in identifying the
significant groups.
Nevertheless, in terms of computation time, the dimension reduction
brought by stacking added significant benefits (around a factor of 2).
In fact, in the \textit{importance block} without stacking, all the variables of the remaining groups are used to predict those of the group of interest. 
Groups with high cardinality (of variables) are challenging in terms of memory
resources and required computation, suggesting that \textit{internal stacking}
can help to reduce computational burden.
%
Real-world empirical application of \textit{BCPI-DNN} with \textit{internal stacking} for age-prediction from brain imaging and socio-demographic information are summarized in Fig.~\ref{ukbb_app}.
Results in \textbf{(Degree of Significance)} ranked the groups according to
their corresponding level of significance. We choose a conservative significance level of $p = 0.001$ (Dashed line at  $log_{10}(0.0001) = 3$).
Using the stacking approach, we scored the heterogeneous  \textit{brain} and \textit{social} input variables regarding their predictive importance.
As expected, we found that the brain groups - excluding \textit{Brain DMRI MD} - were highly
important for age prediction.
Interestingly, \textit{Lifestyle} and \textit{Education} were among the top predictive variables, conditional on the brain groups, suggesting the presence of complementary information.  
To challenge the plausibility of the selected groups, we investigated prediction performance after excluding non-significant groups.
We used 10-fold cross validation with significance estimation and refitting the reduced model using the training set while scoring with the reduced model on the testing set. 
The reduced model did not perform visibly worse than the full model
($R^2=0.8,MAE=2.9$), suggesting that our procedure effectively selects predictive groups.
Of note the performance is in line with state-of-the art benchmarks on the UKBB
based on convolutional neural networks ($MAE \sim$ 2-3 years, e.g.,
\citealt{roibuBrainAgesDerived2023, jonssonBrainAgePrediction2019}).
Consequently, results suggest that the proposed approach combined good prediction performance with effective identification of relevant groups of variables.
For additional supporting results, see supplementary materials.

\section{Discussion}
\label{discussion}
In this work, we proposed \textit{BCPI}, a novel and usable framework for computing single- and group-level variable importance.
Our work provides statistical guarantees based on results from
\textit{Conditional Permutation Importance (CPI)}, whereas our implementation supports arbitrary regression and classification models consistent with the scikit-learn API.
We developed our approach beginning from the observation that standard \textit{Permutation Importance {PI}},
represented by the \textit{BPI-DNN} approach, lacks the ability to control
type-I error \cite{williamsonGeneralFrameworkInference2021} with high correlated settings in Fig.~\ref{bench_plot}, despite the high AUC score \cite{miPermutationbasedIdentificationImportant2021}.
We extended these results, theoretically and empirically, to the group setting by proposing \textit{BCPI-DNN}, which is built on top of an expressive DNN model as a base learner.
This recipe led to high AUC scores while maintaining the control of type-I error across different correlation scenarios (Fig.~\ref{bench_plot}).

Inspired by recent applications of model stacking for handling multiple groups
or input domains~\cite{albuMMStackEnsNewDeep2023,
zhouMultimodalFeatureFusion2021,
engemannCombiningMagnetoencephalographyMagnetic2020}, we
proposed \textit{internal stacking} which implements stacking inside the DNN model, hence, avoids separate optimization problems common for stacking pipelines.
This was achieved through extra sub-linear layers building linear summaries for each group of variables.
Our benchmarks suggested that stacking maintained inferential performance of the full model while bringing time benefits (at least up to a factor of 2), especially for groups with high cardinality of variables (Fig.~\ref{compare_plot}).
Moreover, additional analyses of calibration of
\textit{BCPI-DNN} versus \textit{BPI-DNN} suggested that the p-values for \textit{BCPI-DNN} showed a slightly conservative profile for \textit{BCPI-DNN}.
Instead, \textit{BPI-DNN} showed poor calibration, once more underlining the relevance of conditional permutations.

Our empirical investigation of age prediction using the UKBB dataset suggests that the proposed framework facilitates constructing strong predictions models alongside trustworthy insights on the important predictive inputs.
While prediction performance of our model was in line with state-of-the art
results for the UKBB\citet{roibuBrainAgesDerived2023, jonssonBrainAgePrediction2019}), here, we provided a statistically grounded confirmation for the conclusions drawn in \citealt{dadiPopulationModelingMachine2021} who used a less formal approach consistent with the \textit{LOGI} approach. 

Several limitations apply to our work.
\textit{BCPI-DNN} utilizes a DNN model as the base estimator for its high predictive accuracy.
However, when the amount of training data is limited, the network can potentially memorize the training examples instead of learning generalizable patterns and a simpler base learner might be preferable, e.g. a Random Forest.
Additional analyses of computation time for \textit{BCPI-DNN} in situations of low ($5$) versus high ($100$) cardinality showed that the benefit of~\textit{internal stacking} became more pronounced with larger groups of variables.
This is due to the extra training of the added sub-linear layers.
Our work made use of predefined groups, which may not always be available.
Instead,  statistically defined groups could be used e.g. obtained from clustering.
A possible issue might then be that the groups mix heterogeneous variables, which makes their interpretation challenging.
Furthermore, it is important to apply one-hot encoding of categorical variables after clustering.
On the flip side, reliance on predefined groups may lead to poor inference if the group structure does not track variable importance, e.g. if important variables are distributed in all groups.
This topic deserves careful investigation in the future.
Moreover, here we only performed \textit{internal stacking} by applying linear projection on the input data.
It will be interesting to better understand the potential of non-linear
projections.

Finally, additional possible future directions include studying the impact of
missing and low values on the accuracy, also across different group definitions.

\paragraph{Acknowledgement}
This work has been supported by Bertrand Thirion and is supported by the KARAIB
AI chair (ANR-20-CHIA-0025-01),  and the H2020 Research Infrastructures Grant
EBRAIN-Health 101058516.
D.E. is a full-time employee of F. Hoffmann-La Roche Ltd.

\bibliography{Biblio}

\clearpage
\appendix
\section{Conditional Permutation Importance (CPI) Wald statistic asymptotically controls type-I errors: hypotheses, theorem and proof}
\label{proof}
\paragraph{Outline}
The proof relies on the observation that the importance score defined in
\eqref{eq:cpi} is $0$ in the asymptotic regime, where the permutation
procedure becomes a sampling step, under the assumption that the subset of
variables $\mathcal{J}$ is not conditionally associated with $y$.
Then all the proof focuses on the convergence of the finite-sample
estimator to the population one. To study this, we use the framework
developed in \citep{williamsonGeneralFrameworkInference2021}.
Note that the major difference with respect to other contributions
\citep{watsonTestingConditionalIndependence2021} is that the ensuing
inference is no longer conditioned on the estimated learner
$\hat{\mu}$.
Next, we first restate the precise technical conditions under which the
different importance scores considered are asymptotically valid,
i.e. lead to a Wald-type statistic that behaves as a standard normal
under the null hypothesis.

\paragraph{Notations}
Let $\mathcal{F}$ represent the class of functions from which a learner $\mu:
\mathbf{x} \mapsto y$ is sought. \\
Let $P_0$ be the data-generating distribution and $P_n$ is the
empirical data distribution observed after drawing $n$ samples (noted
$n_{train}$ in the main text; in this section, we denote it $n$ to
simplify notations).
The separation between train and test samples is actually only
relevant to alleviate some technical conditions on the class of
learners used.
$\mathcal{M}$ is the general class of distributions from which $P_1, \dots, P_n, P_0$ are drawn.
$\mathcal{R}:= \{c(P_1 - P_2 ) : c \in [0, \infty), P_1 , P_2 \in \mathcal{M} \}$ is the space of finite signed measures generated by $\mathcal{M}$.
Let $l$ be the loss function used to obtain $\mu$.
Given $f \in \mathcal{F}$, $l(f;P_0) = \int l(f(\mathbf{x}),y) P_0(\mathbf{z}) d\mathbf{z}$, where $\mathbf{z}=(\mathbf{x}, y)$.
Let $\mu_0$ denote a population solution to the estimation problem $\mu_0 \in
\text{argmin}_{f \in \mathcal{F}} l(f;P_0)$ and  $\hat{\mu}_n$ a finite sample
estimate $\hat{\mu}_n \in \text{argmin}_{f \in \mathcal{F}} l(f;P_n) =
\frac{1}{n}\sum_{(\mathbf{x}, y) \in P_n}l(f(\mathbf{x}),y)$. \\
Let us denote by $\dot{l} (\mu, P_0 ; h)$ the Gâteaux derivative of $P \mapsto
l(\mu, P)$ at $P_0$ in the direction $h \in \mathcal{R}$, and define the random
function $g_n : \mathbf{z} \mapsto \dot{l}(\hat{\mu}_n , P_0 ; \delta_\mathbf{z}
- P_0 ) - \dot{l} (\mu_0 , P_0 ; \delta_\mathbf{z} - P_0 )$, where
$\delta_\mathbf{z}$ is the degenerate distribution on $\mathbf{z} = (\mathbf{x},
y)$.
\paragraph{Hypotheses}
\begin{itemize}
\item[(A1)](Optimality) there exists some constant  $C > 0$, such that for each sequence $\mu_1, \mu_2,\cdots \in \mathcal{F}$ given that $\|\mu_n-\mu_0\|\rightarrow 0, |l(\mu_n, P_0) - l(\mu_0, P_0) | < C \|\mu_n-\mu_0\|^2_\mathcal{F}$ for each $n$ large enough.
\item[(A2)](Differentiability) there exists some constant $\kappa > 0$ such that for each sequence $\epsilon_1 , \epsilon_2 , \cdots \in \mathbb{R}$ and $h_1 , h_2 , \cdots \in \mathcal{R}$ satisfying $\epsilon_n \rightarrow 0$ and $\|h_n - h_\infty\| \rightarrow 0$, it holds that
    \begin{equation*}
    \hspace*{-1cm}
    \small
    \underset{{\mu\in\mathcal{F}: \|\mu-\mu_0\|_\mathcal{F}<\kappa}}{\text{sup}} \left | \frac{l(\mu, P_0 + \epsilon_n h_n) - l(\mu, P_0)}{\epsilon_n} - \dot{l}(\mu, P_0; h_n) \right | \rightarrow 0.
    \end{equation*}
    \normalsize
\item[(A3)](Continuity of optimization) $\|\mu_{P_0 + \epsilon h}-\mu_0\|_\mathcal{F} = O(\epsilon)$ for each $h \in \mathcal{R}$.
\item[(A4)](Continuity of derivative) $\mu \mapsto \dot{l}(\mu, P_0;h)$ is continuous at $\mu_0$ relative to $\|.\|_\mathcal{F}$ for each  $h \in \mathcal{R}$.
\item[(B1)] (Minimum rate of convergence) $\|\hat{\mu}_n-\mu_0\|_\mathcal{F} = o_P(n^{-1/4})$.
\item[(B2)] (Weak consistency) $\int g_n(\mathbf{z})^2 dP_0(\mathbf{z}) = o_P(1)$.
\item[(B3)] (Limited complexity) there exists some $P_0$-Donsker class $\mathcal{G}_0$ such that $P_0(g_n \in \mathcal{G}_0) \rightarrow 1$.
\end{itemize}

\paragraph{Proposition} (Theorem 1 in \citep{williamsonGeneralFrameworkInference2021})
If the above conditions hold,  $l(\hat{\mu}_n, P_n)$ is an asymptotically linear
estimator of $l(\mu_0, P_0)$ and $l(\hat{\mu}_n, P_n)$ is non-parametric
efficient. \\
Let $P_0^\star$ be the distribution obtained by sampling the $\mathcal{J}^{th}$ coordinates of $\mathbf{x}$ from the conditional distribution of $q_0(\mathbf{x}^{\mathcal{J}}|\mathbf{x^{-\mathcal{J}}})$, obtained after marginalizing over $y$:
\begin{equation*}
    q_0(\mathbf{x}^{\mathcal{J}}|\mathbf{x^{-\mathcal{J}}}) = \frac{\int P_0(\mathbf{x}, y) dy}{\int P_0(\mathbf{x}, y) d\mathbf{x^\mathcal{J}} dy}
\end{equation*}
$P_0^\star(\mathbf{x}, y) = q_0(\mathbf{x}^{\mathcal{J}}|\mathbf{x^{-\mathcal{J}}}) \int P_0(\mathbf{x}, y) d\mathbf{x^{\mathcal{J}}}$.
Similarly, let $P_n^\star$ denote its finite-sample counterpart.
It turns out from the definition of $\hat{m}^{\mathcal{J}}_{CPI}$ in
Eq. \ref{eq:cpi} that $\hat{m}^{\mathcal{J}}_{CPI}=l(\hat{\mu}_{n},P_n^\star) -
l(\hat{\mu}_n,P_n)$.  It is thus the final-sample estimator of the
population quantity $m^{\mathcal{J}}_{CPI}=l(\hat{\mu}_{0},P_0^\star) -
l(\hat{\mu}_0,P_0)$. \\
Given that $\hat{m}^{\mathcal{J}}_{CPI} = l(\hat{\mu}_{n}, P_n^\star) -
l(\hat{\mu}_{0}, P_0^\star) -\left( l(\hat{\mu}_n,P_n) - l(\hat{\mu}_0,P_0)
\right) + l(\hat{\mu}_{0},P_0^\star) - l(\hat{\mu}_0,P_0)$,
the estimator $\hat{m}^{\mathcal{J}}_{CPI}$ is asymptotically linear and non-parametric efficient.\\
The crucial observation is that under the $\mathcal{J}$-null hypothesis, $y$ is
independent of $\mathbf{x}^{\mathcal{J}}$ given $\mathbf{x^{-\mathcal{J}}}$.
Indeed, in that case $P_0(\mathbf{x},y) = q_0(\mathbf{x}^{\mathcal{J}}|\mathbf{x^{-\mathcal{J}}}) P_0(y|\mathbf{x^{-\mathcal{J}}}) P_0(\mathbf{x^{-\mathcal{J}}})$ and $P_0(\mathbf{x}^{\mathcal{J}}|\mathbf{x^{-\mathcal{J}}}, y) = P_0(\mathbf{x}^{\mathcal{J}}|\mathbf{x^{-\mathcal{J}}})$, so that $P_0^\star = P_0$.
Hence, mean/variance  of $\hat{m}^{\mathcal{J}}_{CPI}$'s distribution provide valid confidence intervals for $m^{\mathcal{J}}_{CPI}$ and $mean(\hat{m}^{\mathcal{J}}_{CPI}) \underset{n\rightarrow\infty}\rightarrow 0$.
Thus, the Wald statistic $\hat{z}^{\mathcal{J}}_{CPI}$ converges to a standard normal distribution, implying that the ensuing test is valid.\\
In practice, hypothesis (B3), which is likely violated, is avoided by
the use of cross-fitting as discussed in
\citep{williamsonGeneralFrameworkInference2021}: as stated in the main
text, variable importance is evaluated on a set of samples not used for
training.
An interesting impact of the cross-fitting approach is that it reduces the hypotheses to (A1) and (A2), plus the following two:
\begin{itemize}
\item[(B1')] (Minimum rate of convergence) $\|\hat{\mu}_n-\mu_0\|_\mathcal{F} = o_P(n^{-1/4})$ on each fold of the sample splitting scheme.
\item[(B2')] (Weak consistency) $\int g_n(\mathbf{z})^2 dP_0(\mathbf{z}) = o_P(1)$ on each fold of the sample splitting scheme.
\end{itemize}

\section{Algorithm for Conditional Permutation Importance (\textit{CPI})}
\begin{algorithm}[ht]
    \begin{algorithmic}[1]
    \REQUIRE $\mathbf{X} \in \mathbb{R}^{n_{test} \times p}$, $\mathbf{y} \in \mathbb{R}^{n_{test}}$, $\hat{\mu}$: estimator, $l$: loss function, RF: learner trained to predict $\mathbf{x}^{\mathcal{J}}$ from $\mathbf{x^{-\mathcal{J}}}$
    \STATE $B \gets$ number of permutations
    \STATE $\mathbf{X^{-\mathcal{J}}} \gets$ $\mathbf{X}$ with
    $\mathcal{J}^{th}$ subset of variables removed 
    \FOR{i = 1 to $n_{test}$} 
        \STATE $\mathbf{\hat{x}^{\mathcal{J}}_i} \gets$ Random Forest($\mathbf{x^{-\mathcal{J}}_i}$)
    \ENDFOR
    \STATE Residuals $\mathbf{\epsilon^{\mathcal{J}}} \gets \mathbf{X^{\mathcal{J}}} - \mathbf{\hat{X}^{\mathcal{J}}}$

    \FOR{b = 1 to B}
    \STATE $\mathbf{\tilde{\epsilon}^{\mathcal{J},b}} \gets$ Joint Random Shuffling($\mathbf{\epsilon^{\mathcal{J}}}$)
    \STATE $\mathbf{\tilde{X}^{\mathcal{J},b}} \gets \mathbf{\hat{X}^{\mathcal{J}}} + \mathbf{\tilde{\epsilon}^{\mathcal{J},b}}$
            \FOR{i = 1 to $n_{test}$} 
            \STATE $\tilde{y}_i^b \gets \hat{\mu}(\mathbf{\tilde{x}^{\mathcal{J}, b}_i})$
                \STATE compute $l_i^{\mathcal{J}, b}$
                \ENDFOR
                \ENDFOR
    \STATE $\text{mean}(\hat{m}_{CPI}^{\mathcal{J}}) = \frac{1}{n_{test}}
    \frac{1}{B} \overset{n_{test}}{\underset{i = 1}{\sum}}
    \overset{B}{\underset{b =
    1}{\sum}} l_i^{\mathcal{J}, b}$ \label{v_imp_mean}
    \STATE $\text{std}(\hat{m}_{CPI}^{\mathcal{J}}) =
    \sqrt{\frac{1}{n_{test}-1} \overset{n_{test}}{\underset{i = 1}{\sum}} \left(
    \frac{1}{B} \overset{B}{\underset{b = 1}{\sum}} l_i^{\mathcal{J}, b} -
    mean(\hat{m}_{CPI}^{\mathcal{J}})\right)^2}$ \label{v_imp_std}
    \STATE $z_{CPI}^{\mathcal{J}} =
    \frac{\text{mean}(\hat{m}_{CPI}^{\mathcal{J}})}{\text{std}(\hat{m}_{CPI}^{\mathcal{J}})}$ \label{z_eq}
    \STATE $p^{\mathcal{J}} \gets  1 - cdf(z^{\mathcal{J}}_{CPI})$ \label{pval}
    \end{algorithmic}
    \caption{\textbf{Conditional sampling step}: The algorithm implements the
    conditional sampling step in place of the permutation approach when
    computing the p-value of group $\mathbf{X^{\mathcal{J}}}$}
    \label{alg_cond}
\end{algorithm}
The loss score $l_i^{\mathcal{J}, b} \in \mathbb{R}$ is defined by: 
\begin{equation*}
    l_i^{\mathcal{J}, b} = 
    \left\{
        \begin{array}{ll}
            y_{i} \log \left(\frac{S(\hat{y}_{i})}{S(\tilde{y}_{i}^{b})} \right) + (1 - y_{i}) \log \left(\frac{1 - S(\hat{y}_{i})}{1 - S(\tilde{y}_{i}^{b})}\right) \\
            (y_{i} - \tilde{y}_{i}^{b})^{2} - (y_{i} - \hat{y}_{i})^{2}
        \end{array}
    \right.
\end{equation*}
for binary and regression cases respectively where $i \in \llbracket n_{test}
\rrbracket$, $\mathcal{J} \in (\mathcal{S} \cup \mathcal{S'})$, $b \in
\llbracket B \rrbracket$, $\hat{y}_{i} = \hat{\mu}(\mathbf{x_{i}})$ and
$\tilde{y}_{i}^{b} = \hat{\mu}(\mathbf{\tilde{x}_{i}^{\mathcal{J}, b}})$ is the newly
predicted value following the reconstruction of the group of interest with
$b^{th}$ residual shuffled and $S(x)=\frac{1}{1+e^{-x}}$. 

\section{Evaluation Metrics}
\label{metrics}
\paragraph{AUC score} \citep{bradleyUseAreaROC1997}: 
The variables are ordered by increasing p-values, yielding a family of
$p$ splits into relevant and non-relevant at various thresholds.
AUC score measures the consistency of this ranking with the ground
truth ($n_{signals}$ predictive features versus $p-n_{signals}$).
\paragraph{Type-I error}: Some methods output p-values for each of the
variables, that measure the evidence against each variable being a null
variable.
This score checks whether the rate of low p-values of null variables
is not exceeding the nominal false positive rate (set to 0.05).
\paragraph{Power}: This score reports the average proportion of informative variables detected (when considering variables with p-value $< 0.05$).
\paragraph{Computation time}: The average computation time per core on 100 cores.
\paragraph{Prediction Scores}: As some methods share the same core to perform
inference and with the data divided into a train/test scheme, we evaluate the
predictive power for the different cores on the test set.

\section{Pre-defined groups in UK BioBank}
\label{table_groups}

\begin{table}[ht]
    \centering
    
    \begin{tabular}{l l l}
        \textbf{Index} & \textbf{Name} & \textbf{\# variables} \\ 
        \hline \\
        1 & Connectivity (FMRI) & 1485 \\
        2 & Brain DMRI FA & 48 \\
        3 & Brain DMRI ICVF & 48 \\
        4 & Brain DMRI ISOVF & 48 \\
        5 & Brain DMRI L1 & 48 \\
        6 & Brain DMRI L2 & 48 \\
        7 & Brain DMRI L3 & 48 \\
        8 & Brain DMRI MD & 48 \\
        9 & Brain DMRI MO & 48 \\
        10 & Brain DMRI OD & 48 \\
        11 & Brain SMRI & 157 \\
        12 & Early-Life & 8 \\
        13 & Education & 2 \\
        14 & Lifestyle & 45 \\
        15 & Mental Health & 25 \\
        16 & Demographics & 1
    \end{tabular}

    \caption{\textbf{Knowledge-based groups in UK BioBank}: Imaging and
    socio-demographic formed groups within the data from UK Biobank with their
    corresponding cardinalities. \textit{FMRI}: Functional Magnetic Resonance
    Imaging. 
    Following \cite{taeCurrentClinicalApplications2018, chenImprovingEstimationFiber2016}, \textit{DMRI}:
    Diffusion Magnetic Resonance Imaging,
    \textit{FA}: Fractional anisotropy (a measure of the degree of anisotropy of water diffusion in tissue),
    \textit{ICVF}: IntraCellular Volume Fraction (a measure of the amount of
    space in tissue occupied by intracellular water),
    \textit{ISOVF}: ISOtropic Volume Fraction (a measure of the amount of
    space in tissue occupied by freely diffusing water),
    \textit{L1}: The largest eigenvalue of the diffusion tensor and indicates
    the rate of diffusion in the direction of the greatest diffusion,
    \textit{L2}: An intermediate in size eigenvalue of the diffusion tensor and
    indicates the rate of diffusion in the direction perpendicular to the
    direction of the greatest diffusion,
    \textit{L3}: The smallest eigenvalue of the diffusion tensor and indicates
    the rate of diffusion in the direction perpendicular to the first two
    directions,
    \textit{MD}: Mean Diffusivity (a measure of the average rate of water
    diffusion in all directions),
    \textit{MO}: Mode (a probabilistic tractography measure for crossing white
    matter fibers),
    \textit{OD}: A measure of the angular difference between two sets of
    directions,
    \textit{SMRI}: Structural Magnetic Resonance Imaging.
    }
    \label{table1}
\end{table}

\onecolumn
\section{Calibration of p-values between \textit{BCPI-DNN} and
\textit{BPI-DNN}}
\label{calibration}
\begin{figure}[h]
    \centering
    \includegraphics[width=0.9\textwidth]{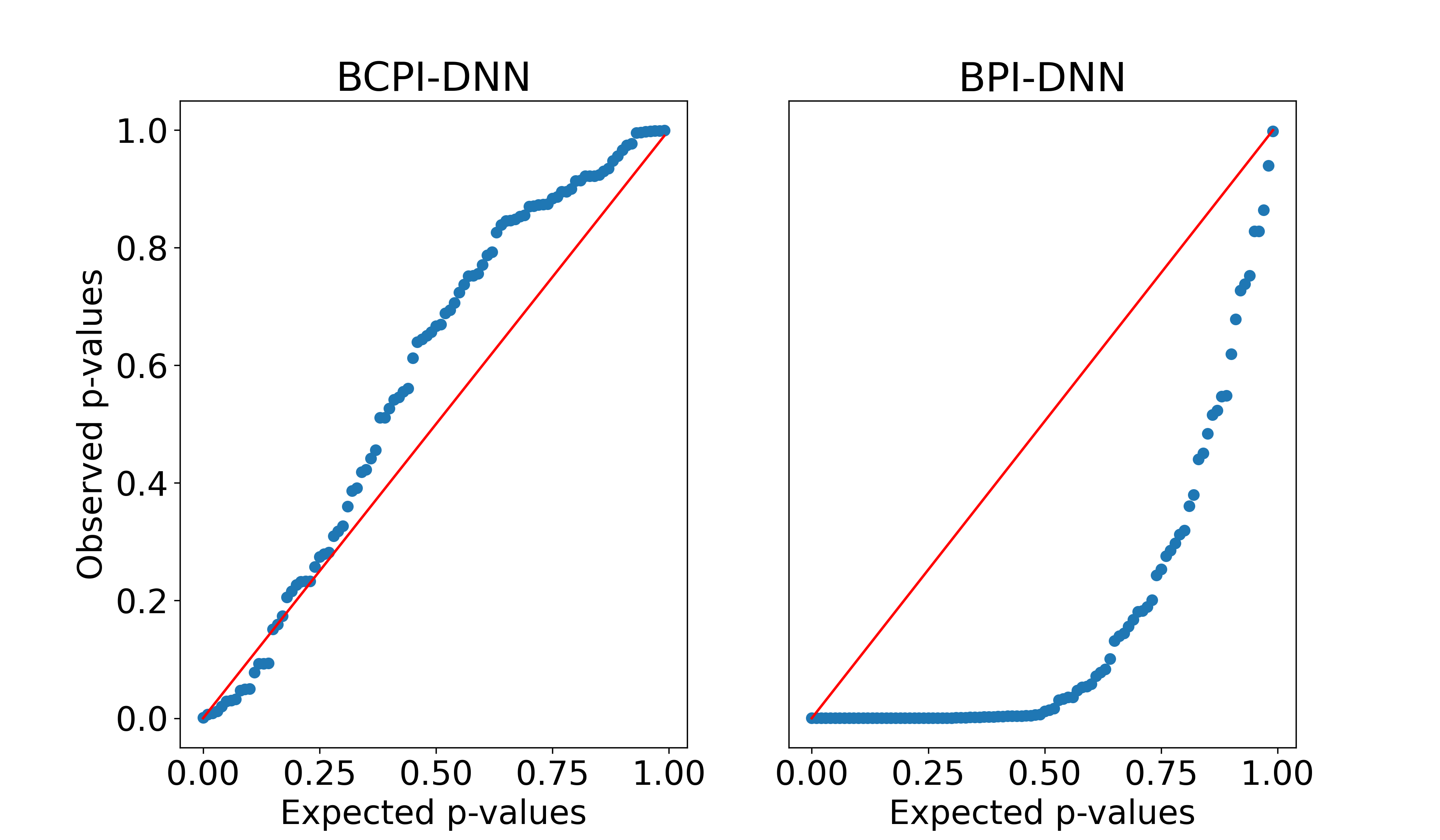}
    \caption{\textbf{p-values calibration}: The calibration of p-values ensuing
    from \textit{BCPI-DNN} with the \textit{conditional permutation} approach is
    compared to that of \textit{BPI-DNN} with \textit{standard permutation}
    approach. The p-value's distribution of one randomly selected non significant
    variable is compared to the uniform distribution. Prediction task was
    simulated with $n$ = $1000$ and $p$ = $50$.}
\end{figure}

\renewcommand{\thefigure}{1 - S1}
\onecolumn
\section{Supplement Figure 1 - Power \& Computation time}
\label{supp_pow_time}
\begin{figure}[h]
    \centering
    \includegraphics[width=0.9\textwidth]{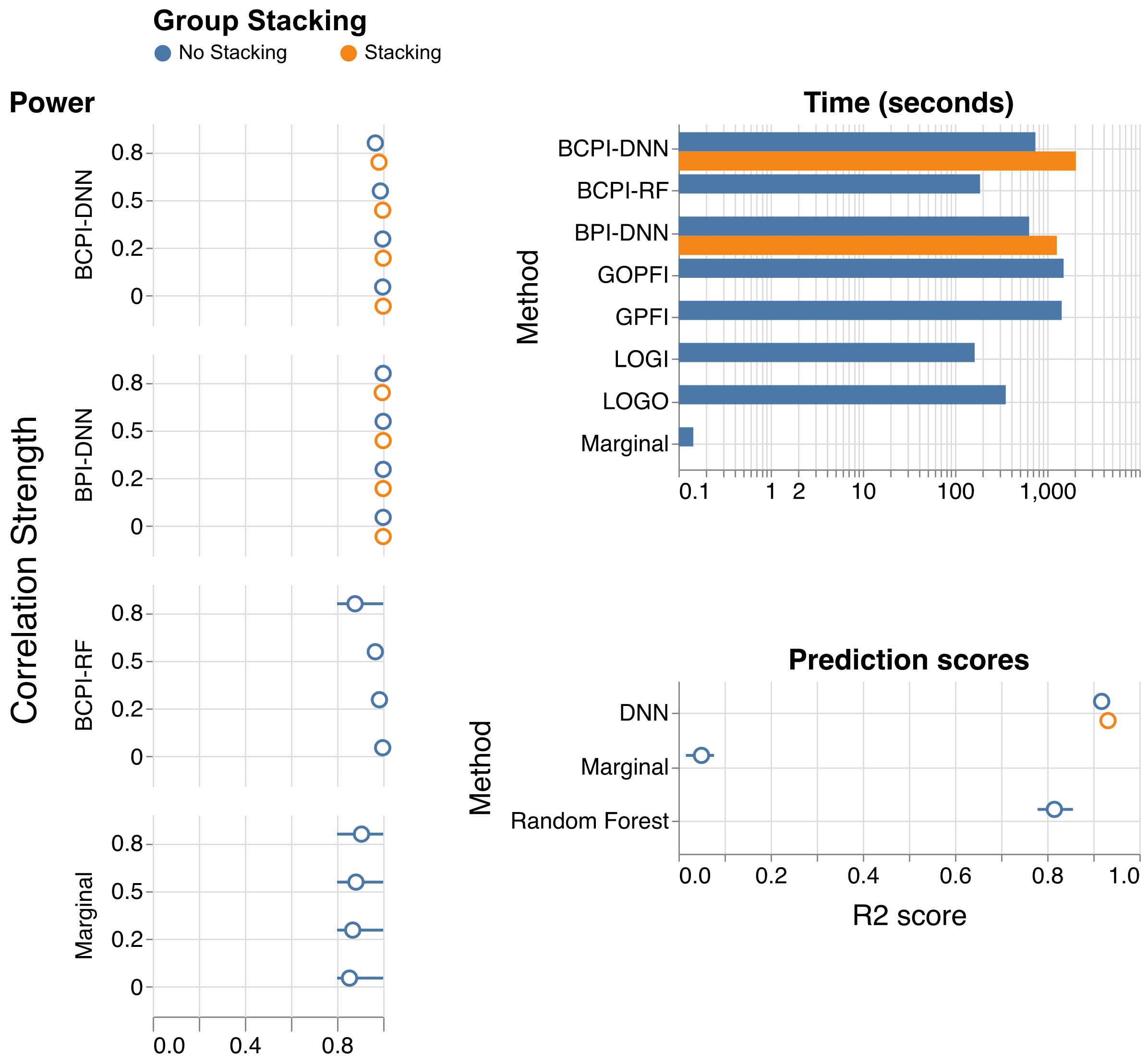}
    \caption{\textbf{Benchmarking grouping methods}: \textit{BCPI-DNN} is
    compared to baseline models and competing approaches for group variable
    importance providing p-values. \textbf{(Power)} indicates the mean
    proportion of informative variables identified. \textbf{(Time)} reports the
    computation time in seconds with $\log_{10}$ scale per core on $100$ cores.
    \textbf{(Prediction scores)} presents the performance of the different base
    learners used in the group variable importance methods (\textit{Marginal}: \{
    Marginal effects\}, \textit{Random Forest}: \{BCPI-RF, LOGI, LOGO, GPFI \&
    GOPFI\}, \textit{DNN}: \{BPI-DNN \& BCPI-DNN\}). Prediction tasks were
    simulated with $n$ = $1000$ and $p$ = $50$.}
\end{figure}

\renewcommand{\thefigure}{1 - S2}
\onecolumn
\section{Supplement Figure 1 - AUC score for \textit{Grouped Shapley} values}
\label{auc_sage}
\begin{figure}[h]
    \centering
    \includegraphics[width=0.3\textwidth]{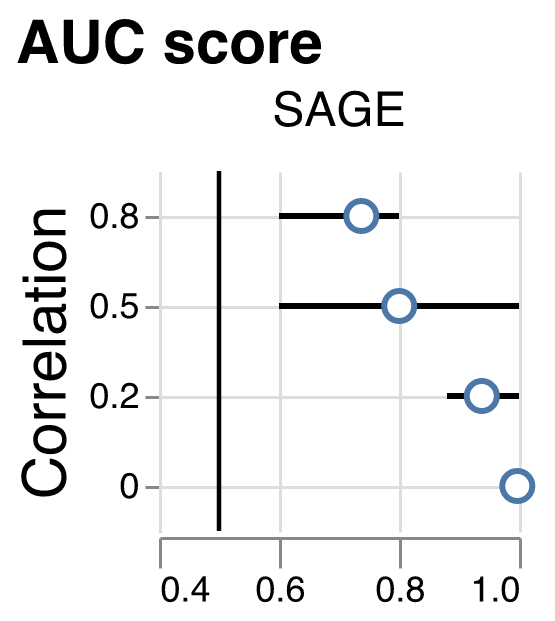}
    \caption{\textbf{\textit{Grouped Shapley} values}: Prediction tasks were
    simulated with $n$ = $1000$ and $p$ = $50$. Solid line: chance level.}
\end{figure}
The grouped version of SAGE (Global Importance with Shapley values
\cite{covertUnderstandingGlobalFeature2020}) was assessed with AUC scores (for
detecting important variables) as it does not provide p-values.
SAGE performed well in low-correlation settings (mean $\approx 0.95$) but the performance
dropped in high-correlation settings (mean $\approx 0.76$).
\twocolumn

\renewcommand{\thefigure}{1 - S3}
\onecolumn
\section{Supplement Figure 1 - AUC score \& Type-I error (Non linear case)}
\label{auc_type1_NonLinear}
\begin{figure}[h]
    \centering
    \includegraphics[width=0.9\textwidth]{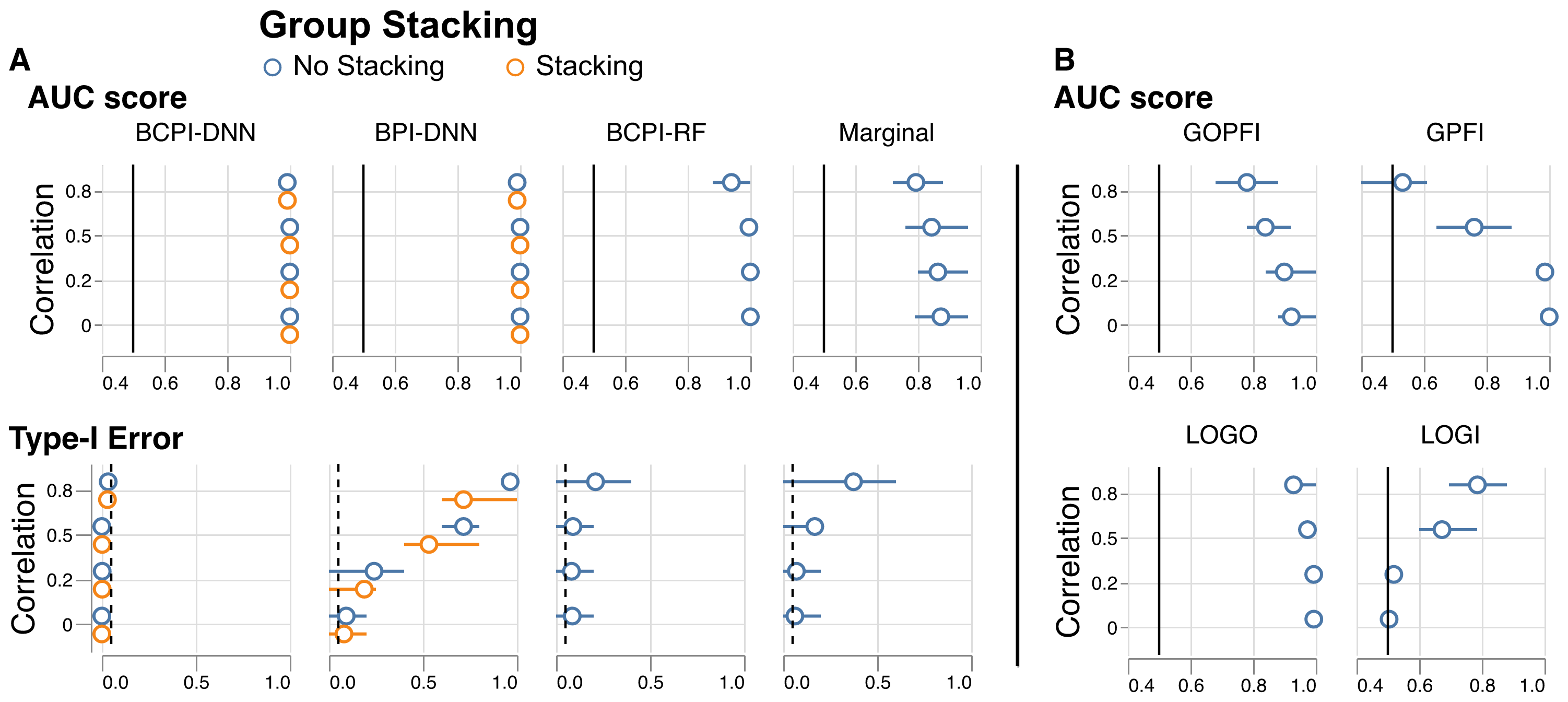}
    \caption{\textbf{Benchmarking grouping methods}: \textit{BCPI-DNN} is
    compared to baseline models and competing approaches for group variable
    importance. It encompasses two panels: \textbf{(A)} for the methods providing
    p-values used to check for AUC score and for statistical guarantees (Type-I
    error control), and \textbf{(B)} for the methods deprived of p-values, thus
    the importance scores are used to check for AUC score. Prediction tasks were
    simulated with $n$ = $1000$ and $p$ = $50$. Dashed line: targeted type-I error rate at 5\%. Solid line: chance level.}
\end{figure}
To make the data-generating process more complex, we have added pair interactions to
the regression simulation introduced in Fig.~\ref{bench_plot}.
The new outcome is set to:
$ y_{i} = \mathbf{x_i} \boldsymbol{\beta^{main}} + \text{quad}(\mathbf{x_i,
\beta^{quad}}) + \sigma \epsilon_i,  \; \forall i \in \llbracket n \rrbracket$
where the magnitude $\sigma$ of the noise is set to $\frac{||\mathbf{X}
\boldsymbol{\beta^{\text{main}}} +  \text{quad}(\boldsymbol{X, \beta^{quad}})
||_{2}}{SNR \sqrt{n}}$ and $\textrm{quad}(\boldsymbol{x_i, \beta^{quad}}) = \displaystyle\sum_{\underset{k < j}{k, j=1}}^{p_{signals}} \boldsymbol{\beta^{quad}}_{k, j} x^{k}_i x^{j}_i$.
The results show that \textit{BCPI-DNN} outperforms all the alternatives methods
presenting high AUC performance coupled with a control for type-I error under
the predefined nominal rate.
\textit{BCPI-RF}, where the inference estimator is
a Random Forest, showed an almost similar good performance with a little drop in
high-correlated settings which can be explained by the drop in the predictive
capacity following the plug of the Random Forest.
\twocolumn

\onecolumn
\renewcommand{\thefigure}{1 - S4}
\section{Supplement Figure 1 - Power \& Computation time (Non linear case)}
\label{supp_pow_time_NonLinear}
\begin{figure}[h]
    \centering
    \includegraphics[width=0.9\textwidth]{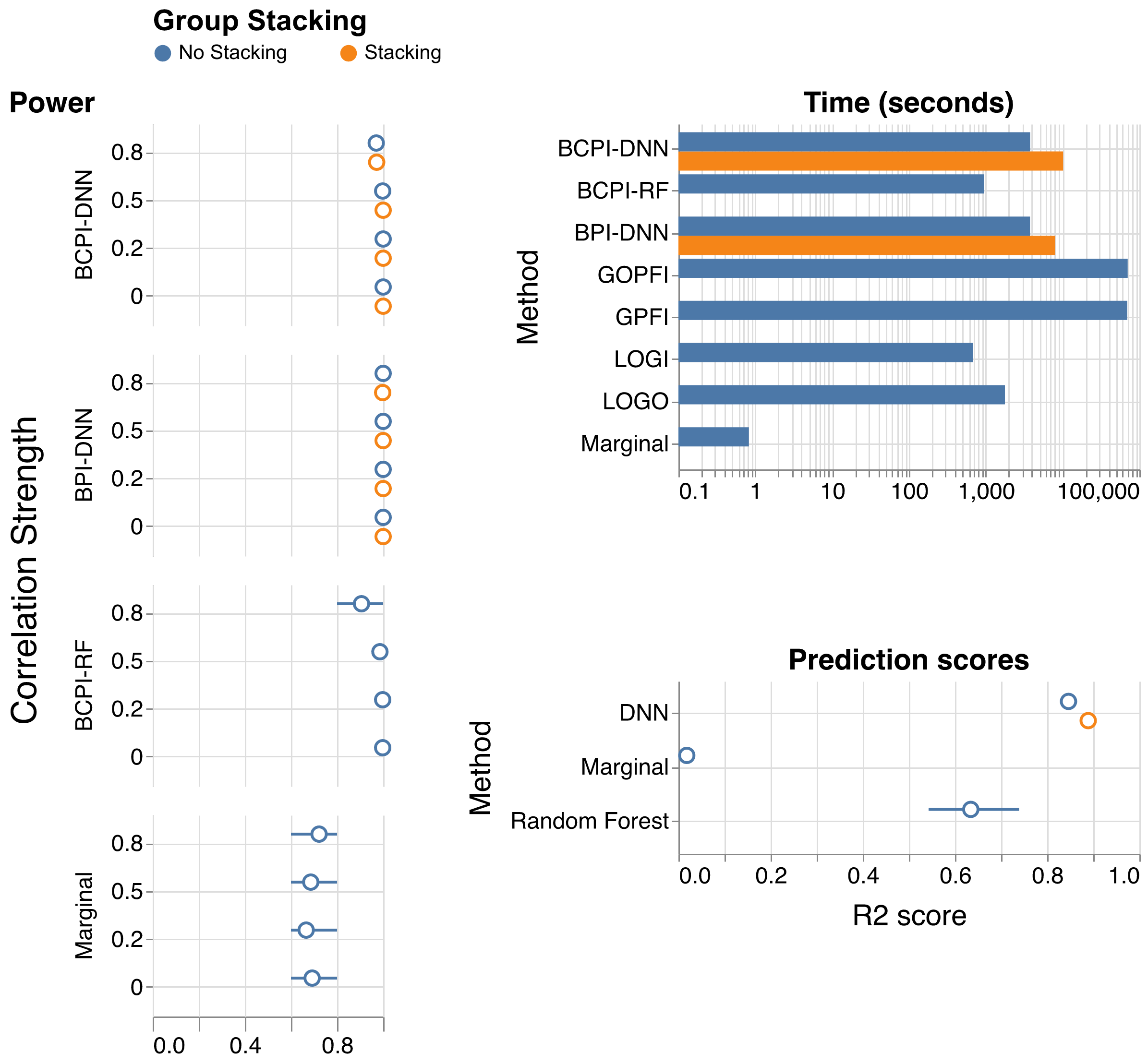}
    \caption{\textbf{Benchmarking grouping methods}: \textit{BCPI-DNN} is
    compared to baseline models and competing approaches for group variable
    importance providing p-values. \textbf{(Power)} indicates the mean
    proportion of informative variables identified. \textbf{(Time)} reports the
    computation time in seconds with $\log_{10}$ scale per core on $100$ cores.
    \textbf{(Prediction scores)} presents the performance of the different base
    learners used in the group variable importance methods (\textit{Marginal}: \{
    Marginal effects\}, \textit{Random Forest}: \{BCPI-RF, LOGI, LOGO, GPFI \&
    GOPFI\}, \textit{DNN}: \{BPI-DNN \& BCPI-DNN\}). Prediction tasks were
    simulated with $n$ = $1000$ and $p$ = $50$.}
\end{figure}
The results showed that \textit{BCPI-DNN}, \textit{BPI-DNN}, \textit{BCPI-RF}
and \textit{Marginal} attained a high performance.
\twocolumn

\onecolumn
\renewcommand{\thefigure}{2 - S1}
\section{Supplement Figure 2 - Groups with different cardinalities}
\label{supp_groups_card}
\begin{figure}[h]
    \centering
    \includegraphics[width=0.9\textwidth]{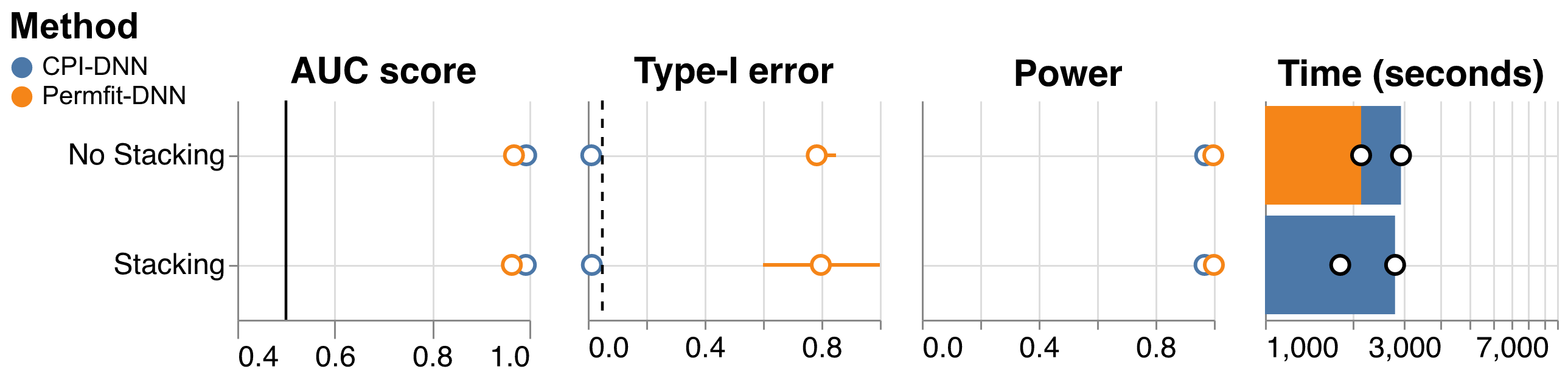}
    \caption{\textbf{Groups of different cardinalities}: The performance of
    \textit{BCPI-DNN} and \textit{Permfit-DNN} at detecting
    important groups on simulated data with $n$ = $1000$ and $p$ = $1000$ with
    $10$ blocks/groups, each group having a cardinality of $10$ with or without
    the \textit{stacking} approach. The \textbf{(AUC
    score)} evaluates the extent to which variables are ranked consistently with
    the ground truth. The \textbf{(Type-I error)} assesses the rate of low
    p-values ($\text{p-val} < 0.05$). \textbf{(Power)} provides information on
    the average proportion of detected informative variables ($\text{p-value} <
    0.05$). The \textbf{(Time)} panel displays computation time in seconds with
    $\log_{10}$ scale per core on $100$ cores. Dashed line: targeted type-I error
    rate. Solid line: chance level.}
\end{figure}
The results showed that \textit{BCPI-DNN}'s capacity to achieve high AUC
performance coupled with a control of Type-I error under the predefined nominal
rate was maintained while providing groups of different cardinalities.
\twocolumn

\end{document}